%% file: main.tex
\documentclass[letterpaper, 10 pt, conference]{ieeeconf}  

\IEEEoverridecommandlockouts                              

\usepackage{graphicx} 
\usepackage{subfig}
\usepackage{amsmath} 
\usepackage{amssymb}  
\usepackage{lipsum}
\usepackage{csquotes}
\usepackage{xcolor}
\usepackage{svg}

\title{\LARGE \bf
iMHS: An Incremental Multi-Hypothesis Smoother}

\author{
Fan Jiang, Varun Agrawal, Russell Buchanan, Maurice Fallon, and Frank Dellaert%
\thanks{Institute for Robotics and Intelligent Machines, College of Computing,
Georgia Institute of Technology,
Atlanta, GA, \{fan.jiang,varunagrawal,frank.dellaert\}@gatech.edu\newline 
Oxford Robotics Institute, Department of Engineering Science,
University of Oxford,
Oxford, UK, \{mfallon,russell\}@robots.ox.ac.uk}%
\thanks{The NASA University Leadership Initiative (grant \#80NSSC20M0163) provided funds to assist the authors with their research, but this article solely reflects the opinions and conclusions of its authors and not any NASA entity.}
}

\begin{document}

\maketitle
\thispagestyle{empty}
\pagestyle{empty}

\begin{abstract}

\input{0_abstract}

\end{abstract}

\section{INTRODUCTION}
\input{1_intro}

\section{BACKGROUND}
\input{2_background}

\section{APPROACH}
\input{3_approach}

\section{EXPERIMENTS}
\input{5_experiments}

\section{CONCLUSIONS}
\input{6_conclusions}




\bibliographystyle{abbrv}
\bibliography{MHT,other}

\end{document}

%% file: 0_abstract.tex
State estimation of multi-modal hybrid systems is an important problem with many applications in the field robotics. 
However, incorporating discrete modes in the estimation process is hampered by a potentially combinatorial growth in computation. 
In this paper we present a novel \textit{incremental} multi-hypothesis smoother based on eliminating a hybrid factor graph into a multi-hypothesis Bayes tree, which represents possible discrete state sequence hypotheses.
Following iSAM, we enable incremental inference by conditioning the past on the future but we add to that the capability of maintaining multiple discrete mode histories, exploiting the temporal structure of the problem to obtain a simplified representation that unifies the multiple hypothesis tree with the Bayes tree. 
In the results section we demonstrate the generality of the algorithm with examples in three problem domains: lane change detection (1D), aircraft maneuver detection (2D), and contact detection in legged robots (3D).

%% file: 1_intro.tex
State estimation of multi-modal hybrid systems is an important problem with many compelling applications. A sample of these applications include fault detection~\cite{Eide96taes_mmae, Hanlon00taes_mmae}), air traffic control (ATC)~\cite{Hwang06cta_estimation_hybrid, Yepes07gcd_aircraft_intent}, decoding cortical motor activity~\cite{Wu04rbme_skf}, and contact dynamics in legged robots~\cite{Grizzle15automatica_walking_survey}. In the last example,  accurate determination of contact state in manipulation and/or walking is challenging yet vital when executing accurate manipulation or locomotion plans.

However, incorporating discrete modes in the estimation process suffers from combinatorial growth in computation. Hence, there is a long history of work on filtering-based approaches, the most well-known of which are model adaptive estimation (MMAE)~\cite{Maybeck79book_sme} and 
the interacting multiple model (IMM)~\cite{BarShalom01book_tracking}. This line of work has been further expanded to switching dynamics Bayes nets by Murphy~\cite{Murphy88tr_switching_kf, Murphy02thesis_dbn}.
Optimizing over the entire time history or \textit{smoothing} is much more challenging, although it has been approximated using Markov Chain Monte Carlo (MCMC) sampling~\cite{Oh05aaai_ddmcmc,Oh05iccv_parametric} and variational inference~\cite{Dong20icml_switching}.
Various methods which keep track of sets of hypotheses has been proposed in the multi-target tracking literature, including the now classical multiple hypothesis tracking (MHT) filter~\cite{Reid79tac_MHT,Cox96pami_efficient_MHT}, the probabilistic MHT~\cite{Streit95tech_PMHT}, JPDAF~\cite{BarShalom01book_tracking}. Both particle-filtering~\cite{Vermaak05itae_MC_JPDAF} and MCMC smoothing~\cite{Khan05pami_mcmc_pf, Khan06pami_mcmc_da, Oh09itac_mcmcda} have been applied.

In this paper we present a novel \textit{incremental} multi-hypothesis smoother based on eliminating a hybrid factor graph into a multi-hypothesis Bayes tree, which represents possible discrete state sequence hypotheses.
We base our work on iSAM~\cite{Kaess08tro_iSAM, Kaess12ijrr_isam2}, which enables incremental inference by conditioning the past on the future, but we add to that the capability of maintaining multiple discrete mode histories.
In contrast to sampling-based and variational approaches, the computational cost is bounded by meaningful thresholds on the probability of state histories, and incremental inference is constant-time.
While our approach is similar in spirit to MH-iSAM2 by Hsiao and Kaess~\cite{Hsiao19icra_mh_iSAM}, 
we exploit the temporal structure of the problem to obtain a simplified representation that unifies the multiple hypothesis tree with the Bayes tree. 
In addition, we support stochastic transitions over the mode sequence, and can properly handle model selection when  modes are not all equally informative about the dynamics or measurement models.

\begin{figure}
\centering
\subfloat[Factor graph for hybrid switching system, with continuous states $x_k$ evolving according to discrete modes $m_k$.]{\includegraphics[width=0.70\linewidth]{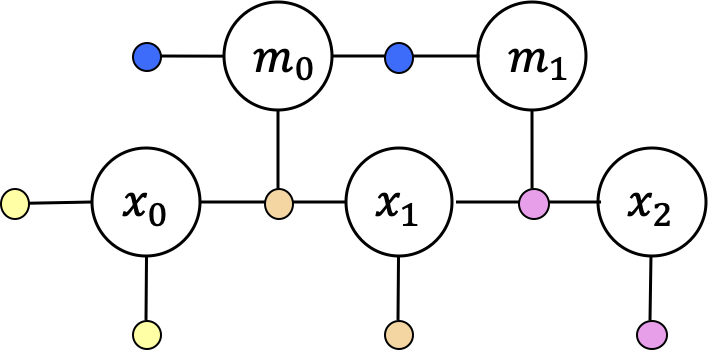}\label{fig:toy-fg}}%
\hfill%
\subfloat[The proposed multi-hypothesis smoother conditions the past on the future in parallel branches of a tree.]{\includegraphics[width=0.60\linewidth]{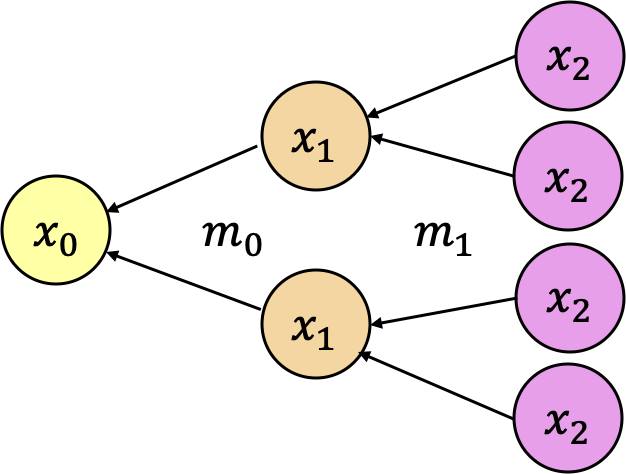}\label{fig-toy-mhs}}
\caption{The iMHS concept illustrated for a toy example.}
\label{fig:toy}
\end{figure}

A small example in Figure \ref{fig:toy} demonstrates the basic idea. The batch problem is represented as a factor graph with the state at three time instants, and a Markov chain of discrete modes that select a different process model at each time step. The unary factors are measurements and will help determine which process model is most likely. 
In the case of a binary mode variable, the mode selects between two process models. For this example with $K=3$ states there are eight possible discrete mode sequences. A naive algorithm would eliminate the factor graph separately for each sequence. However, Figure \ref{fig-toy-mhs} shows that the eliminated factor graphs
can be combined into a multiple-hypothesis tree of conditionals, conditioned on the future states. 

In the results section we show examples from vehicle lane-change detection, aircraft tracking, and contact estimation of legged robots, all of which can be modeled in the above framework.
In general, any system described by a dynamic Bayes net with a combination of discrete and continuous variables can be accommodated.

In summary, the contributions of this work are:
\begin{itemize}
    \item A novel, incremental smoother algorithm for hybrid/multi-modal systems.
    \item A novel data structure that unifies multiple hypothesis trees and Bayes trees.
    \item A demonstration of the generality of the algorithm with examples in three problem domains: lane change detection (1D), aircraft maneuver detection (2D), and contact detection in legged robots (3D).
\end{itemize}

%% file: 2_background.tex
In this paper we rely heavily on the use of graphical models, including dynamic Bayes nets, factor graphs, and incremental inference, which we will review in this section.

\subsection{Problem statement and notation}
\newcommand{\XX}{\mathcal{X}}
\newcommand{\ZZ}{\mathcal{Z}}
\newcommand{\MM}{\mathcal{M}}
\newcommand{\States}{X^K}
\newcommand{\Modes}{M^{K-1}}
\newcommand{\Measurements}{Z^K}
\newcommand{\lastx}{x_{K-1}}
\newcommand{\nextx}{x_K}
\def\MAP{^*}

\begin{figure}
\centering
\subfloat[][
Our notation and independence assumptions are summarized in the dynamic Bayes net above, here for $K=4$.
]{
\includegraphics[width=0.75\linewidth]{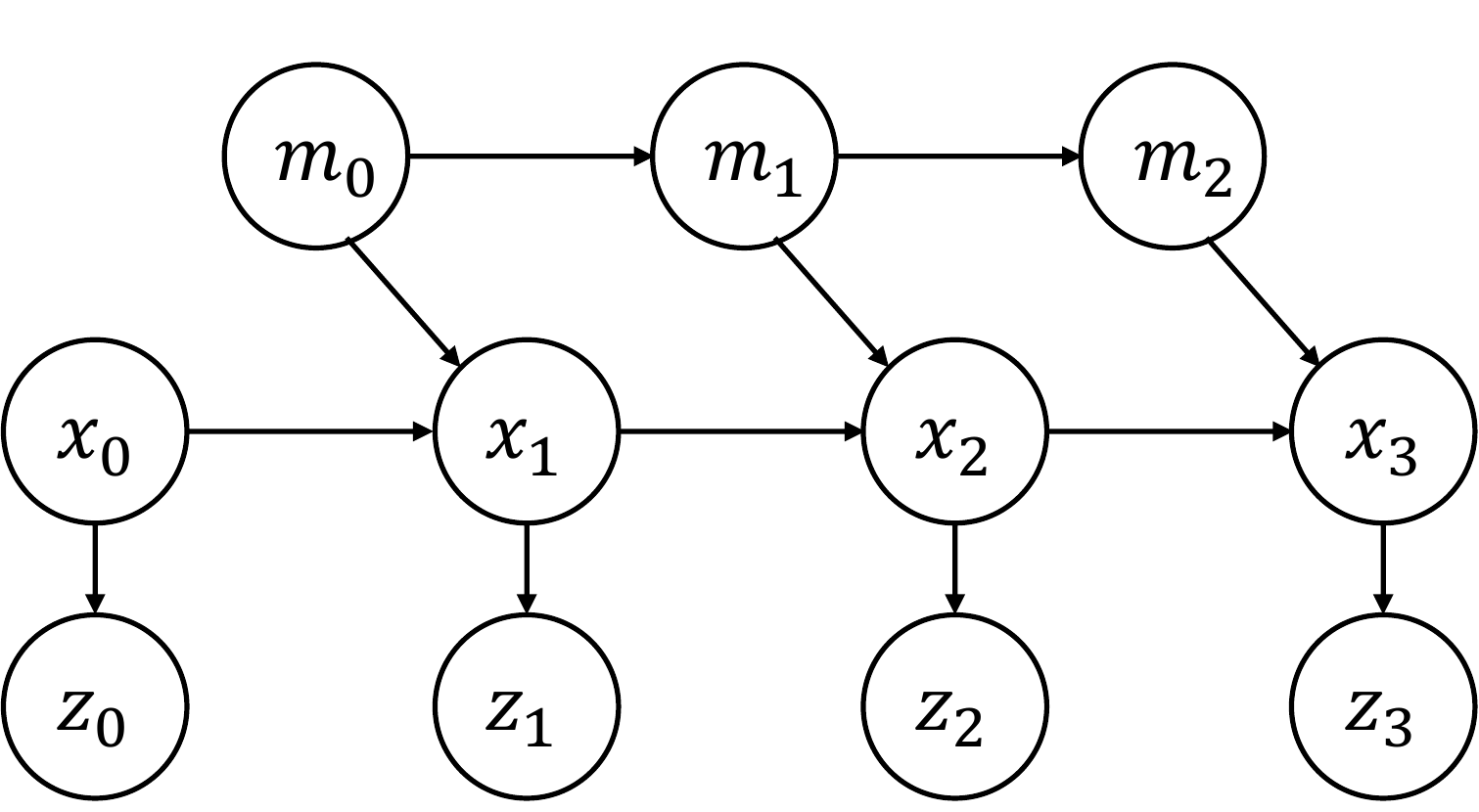}\label{fig:dbn}
}

\subfloat[][
If the modes $\Modes$ are \textit{known}, smoothing is done using the factor graph above, derived from Figure \ref{fig:dbn}.
]{
\includegraphics[width=0.9\linewidth]{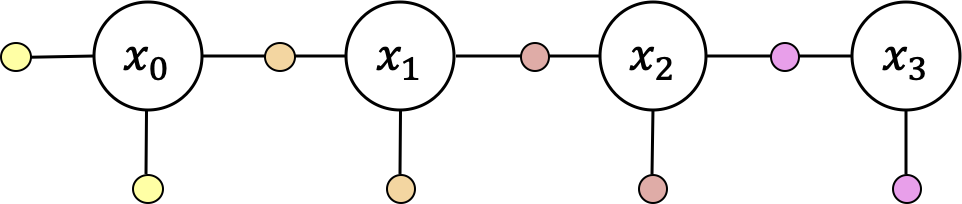}\label{fig:fix-fg}
}
\caption{Notation and independence assumptions.}
\label{fig:modeling}
\end{figure}

Continuous smoothing is the problem of recovering the posterior density
\begin{align}
    p(\States|\Modes, \Measurements),\label{eq:posterior}
\end{align} where $\States \doteq \{x_k\}_{k=0}^{K-1}$ is a sequence of $K$ states from time $t_0$ up to and including time $t_{K-1}$, with $x_k\in \XX$ representing the state at time $t_k$. Given a similarly defined sequence of discrete modes $\Modes$ and measurements $\Measurements$, with $m_k\in\MM$ and $z_k\in\ZZ$, we model the joint density over all variables using the following dynamic Bayes net~\cite{Murphy02thesis_dbn} (Figure \ref{fig:dbn}):
\begin{align}
    p(\States,\Modes,&\Measurements) = P(x_0)p(z_0|x_0)p(m_0) \nonumber \\
    &\times \prod_{k=1}^{K-1} p(m_k,x_k,z_k|m_{k-1},x_{k-1}).\label{eq:dbn}
\end{align}
We assume the mode sequence $\Modes$ is modeled by a Markov chain $P(m_k|m_{k-1})$.
The continuous states evolve via a switching motion model $p(x_k|x_{k-1}, m_{k-1})$,
and measurements are modeled on the continuous states as $p(z_k|x_k)$. One time slice in the dynamic Bayes net is given by
\begin{align}
    &p(m_k,x_k,z_k|m_{k-1},x_{k-1}) \nonumber\\
    &= P(m_k|m_{k-1}) p(x_k|x_{k-1}, m_{k-1}) p(z_k|x_k).
\end{align}

\subsection{Factor Graphs and Trajectory Optimization}

When both discrete modes $\Modes$ and measurements $\Measurements$ are known, it is convenient to represent the posterior density \eqref{eq:posterior} using a factor graph~\cite{Dellaert17fnt_FactorGraphs} as illustrated in Figure \ref{fig:fix-fg},
\begin{align}
    \phi(\States) &\propto p(\States|\Modes,\Measurements) \nonumber \\
    & \doteq \phi(x_0)\phi^z_0(x_0)\prod_{k=1}^{K-1} \phi^m_k(x_{k-1},x_k) \phi^z_k(x_k)
    \label{eq:fg1}
\end{align}
with the factors defined in correspondence with \eqref{eq:dbn}.

The maximum a posteriori (MAP) estimate can be recovered by maximizing $\phi(\States)$ with respect to the continuous states $\States$, also known as trajectory optimization:
\begin{align}
    \States{\MAP} = \arg \min_{\States} \phi(\States).
\end{align}
By fitting a quadratic to the log-posterior, e.g., using the estimate $A^T A$ to the Fisher information matrix $\mathcal{I}$, where 
\begin{align}
    A \doteq - \frac{ \partial \log \phi(\States)} {\partial \States}, \label{eq:Jacobian}
\end{align}
we can obtain an approximation to the posterior as
\begin{align}
    p(\States|\Modes,\Measurements) \approx \mathcal{N}(\States; \States{\MAP}, \mathcal{I}^{-1}), \label{eq:Fisher}
\end{align}
where $\mathcal{N}(x; \mu, \Sigma)\propto \exp \left\{-0.5\|x-\mu\|^2_\Sigma\right\}$ is the multivariate normal distribution with mean $\mu$ and covariance $\Sigma$.

\subsection{The linear case: Kalman smoothing}
In the common case when all continuous densities are linear-Gaussian and (w.l.o.g.) time invariant, we have
\begin{align}
    &p(x_k|x_{k-1}, m) = \mathcal{N}(x_k; F_m x_{k-1} + B_m u_{k-1}, Q_m)\\
    &p(z|x, m) =\mathcal{N}(z; H_m x, R_m)
\end{align}
with a vector-valued state $\States\in R^n$, a control vector $u^k\in R^p$, and a vector-valued measurement $z_k\in R^m$.
Matrices $F_m$, $B_m$, and $H_m$, as well as the process and noise covariances $Q_m$ and $R_m$
can depend on the mode $m$.
It is convenient to define factors in negative log-space, and to minimize the following least-squares criterion, defined as an additive factor graph:
\begin{align}
    &f(\States) \doteq -\log \phi(\States) \nonumber \\
    & = f^m_0(x_0)+ \sum_{k=1}^{K-1} f^m_k(x_{k-1},x_k) + \sum_{k=0}^{K-1} f^z_k(x_k)
\end{align}
with quadratic factors $f(.) \doteq -\log \phi(.)$ defined as
\begin{align}
    &f^m_1(x_0) = \|x_0 - \mu\|^2_P\\
    &f^m_k(x_{k-1},x_k) = \|x_k - (F_m x_{k-1} + B_m u_{k-1})\|^2_{Q_m} \label{eq:f_motion} \\
    &f^z_k(x_k) = \|H_m x_k - z_k\|^2_{R_m}. \label{eq:f_measurement}
\end{align}
In this case the approximation \eqref{eq:Fisher} is exact, with $A$ the sparse Jacobian of $f(\States)$, and the mean $\States{\MAP}$. These can be found in closed form by solving the normal equations
\begin{align}
(A^T A) \States{\MAP} = A^T b
\end{align}
with the right-hand size (RHS) $b$ suitably defined.

\subsection{Incremental Inference}\label{sec:incremental}
\begin{figure}
    \centering
    \includegraphics[width=0.8\linewidth]{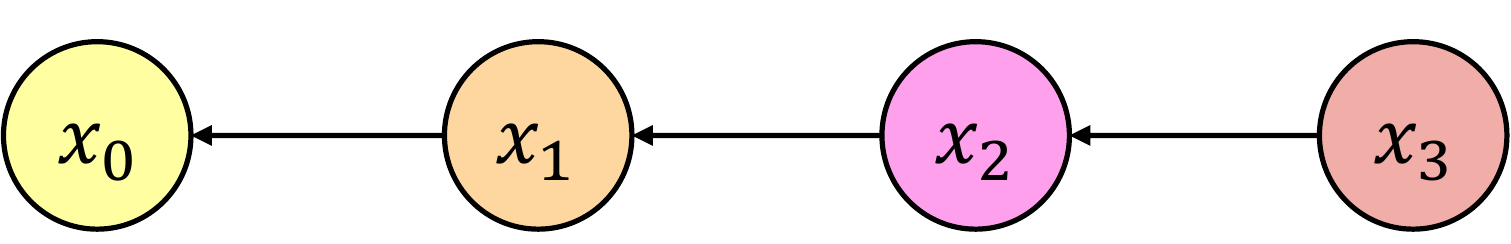}
    \caption{Bayes net representation of posterior $p{\MAP}(\States)$, conditioning the past on the future to enable incremental inference.}
    \label{fig:fix-bn}
\end{figure}

\def\inc{\mathrm{inc}}
Recovering the posterior $p{\MAP}(\States)$ -solving the normal equations in the linear case- can be done using sparse elimination~\cite{Dellaert17fnt_FactorGraphs}. If we eliminate with a natural, time-aligned ordering then under our assumptions the posterior density $p{\MAP}(\States)\doteq p(\States|\Modes, \Measurements)$ will have the form of a chain of conditional densities,
\begin{align}
    \label{eq:chain}
    p{\MAP}(\States)
    = \left\{\prod_{k=1}^{K-1} p{\MAP}(x_{k-1}|x_k)\right\} p{\MAP}(\lastx),
\end{align}
where we use $\MAP$ to indicate that these are conditioned on the measurements and mode sequence. As illustrated in Figure \ref{fig:fix-bn}, with this factorization the past is conditioned on the future, which enables a simple incremental inference scheme. 

The goal of incremental inference is to obtain the posterior density $p{\MAP}(X^{K+1})$ at the next time step given the current posterior density $p{\MAP}(\States)$, the new discrete mode $m_{K-1}$, and a new measurement $z_K$:
\begin{align}
    p(X^{K+1}|M^K,Z^{K+1}) \leftarrow p(\States|\Modes,\Measurements)
\end{align}
Following iSAM2~\cite{Kaess12ijrr_isam2} it suffices to assemble the following factor graph fragment
\begin{align}
\phi(\lastx) \phi^m_k(\lastx,\nextx) \phi^z_K(\nextx)
\end{align}
where the first factor $\phi(\lastx)$ corresponds to the density $p{\MAP}(\lastx)$ from \eqref{eq:chain}.
We then eliminate this into a replacement for $p{\MAP}(\lastx)$, obtaining two conditionals:
\begin{align}
    p{\MAP}(\lastx|\nextx) p{\MAP}(\nextx). \label{eq:bn-fragment}
\end{align}
In the linear case, the incremental update can be done in square-root information form using QR factorization \cite{Bierman78automatica_SRIF, Kaess07icra_iSAM}:
\begin{align}
U
\begin{bmatrix}
  \begin{matrix} A_{K-1} & \\ I_n & - F_m \\ & H_m \end{matrix}
  & \vline &
  \begin{matrix} b_{K-1} \\ B_m u_{K-1} \\ z_K \end{matrix}
\end{bmatrix}
\rightarrow
\begin{bmatrix}
  \begin{matrix} R_m & S_m \\ & T_m \\ & \end{matrix}
  & \vline &
  \begin{matrix} u_m \\ v_m \\ e_m \end{matrix}
\end{bmatrix}
\label{eq:QR}
\end{align}
where the left matrix, augmented with its RHS, corresponds to the factor graph fragment, and 
$U\doteq \mathop{diag}(P,Q_m,R)^{-0.5}$ collects the noise covariances.
The upper-triangular matrix on right, also with RHS, corresponds to the resulting Bayes net fragment~\eqref{eq:bn-fragment}.
Extracting the latter yields the two Gaussian conditional densities on $X_K$ and $\nextx$:
\begin{align}
    p{\MAP}(\lastx|\nextx) &= \mathcal{N}(\lastx; R_m^{-1}(u_m - S_m \nextx), R_m^{-1} R_m^{-T}) \nonumber \\
    p{\MAP}(\nextx) &= \mathcal{N}(\nextx; T_m^{-1}v_m, T_m^{-1} T_m^{-T}).
\end{align}
When the factor graph fragment matrix has rank exactly $2n$, the error at the minimum of the associated quadratic is zero; otherwise, it is $0.5~e_m^T e_m$. We will use this result below.

In the nonlinear case, iSAM~\cite{Kaess08tro_iSAM} and iSAM2~\cite{Kaess12ijrr_isam2} provide incremental management of linearization points, and also handle more general factor graph topologies.

%% file: 3_approach.tex
All of the above assumed known hybrid modes, in which case incremental smoothing is both well known and efficient. In this section we present the main contribution of the paper, which uses a multiple hypothesis framework to effectively run many smoothers in parallel, but arranged in a tree to accommodate shared history. The probability of a particular mode sequence depends both on a Markov chain prior \textit{and} on how compatible the estimated continuous state trajectory is with a particular mode sequence. For example, in a lane change scenario, the changing lateral position of the car provides evidence that a lane change maneuver is in progress.

\subsection{Hybrid State Estimation}

When the hybrid modes $\Modes$ are unknown, we have a hybrid state estimation problem. In this section, we add the mode sequence $\Modes$ as unknowns to the factor graph:
\begin{align}
    &\phi(\States,\Modes) \doteq \phi(m_0)\phi(x_0)\phi^z_0(x_0) \nonumber \\
    &\times \prod_{k=1}^{K-1} \phi(m_{k-1},m_k) \phi(x_{k-1},x_k,m_k) \phi^z_k(x_k)
\end{align}
Above we added a prior on $m_0$, a mode transition model $\phi(m_{k-1},m_k)$, and the motion model factors $\phi^m_k(x_{k-1},x_k,m_k)$ are now also a function of the mode $m_k$.

Within a batch smoother, when we eliminate the states $\States$ with the oldest time-aligned ordering, and \textit{then} eliminate the entire mode sequence, we can obtain a chain of state densities $p(x_{k-1}|x_k, M^k)$ conditioned on the future, and on the mode variables $M^k$ up to that time:
\begin{align}
    &p(\States,\Modes|\Measurements) = \nonumber \\
    &\left\{\prod_{k=1}^{K-1} p\MAP(x_{k-1}|x_k,M^k)\right\} p\MAP(\lastx|\Modes)p\MAP(\Modes)  \label{eq:p_map}
\end{align}
The cardinality of a sequence $M^k$ is $|\MM|^k$, grows exponentially with $k$. Hence, each conditional in the product above is a table of many continuous densities.

Eliminating the last state $\lastx$ yields a new factor $\tau(\Modes)$ on the entire mode sequence:
\begin{align}
    \tau(\Modes) = C_m \int_{\lastx} q\MAP(\lastx) \label{eq:p_modes}
\end{align}
where $q\MAP(\lastx)$ is the unnormalized posterior density on $\lastx$, and $C_m$ is a mode-dependent constant that derives from the motion model and measurement factors in \eqref{eq:fg1}.

\subsection{The Multi-hypothesis Smoother}
\begin{figure}
\centering
\subfloat[][
Eliminated factor graph representing the MHS posterior $p(\States,\Modes|\Measurements)$.
]{
\includegraphics[width=0.8\linewidth]{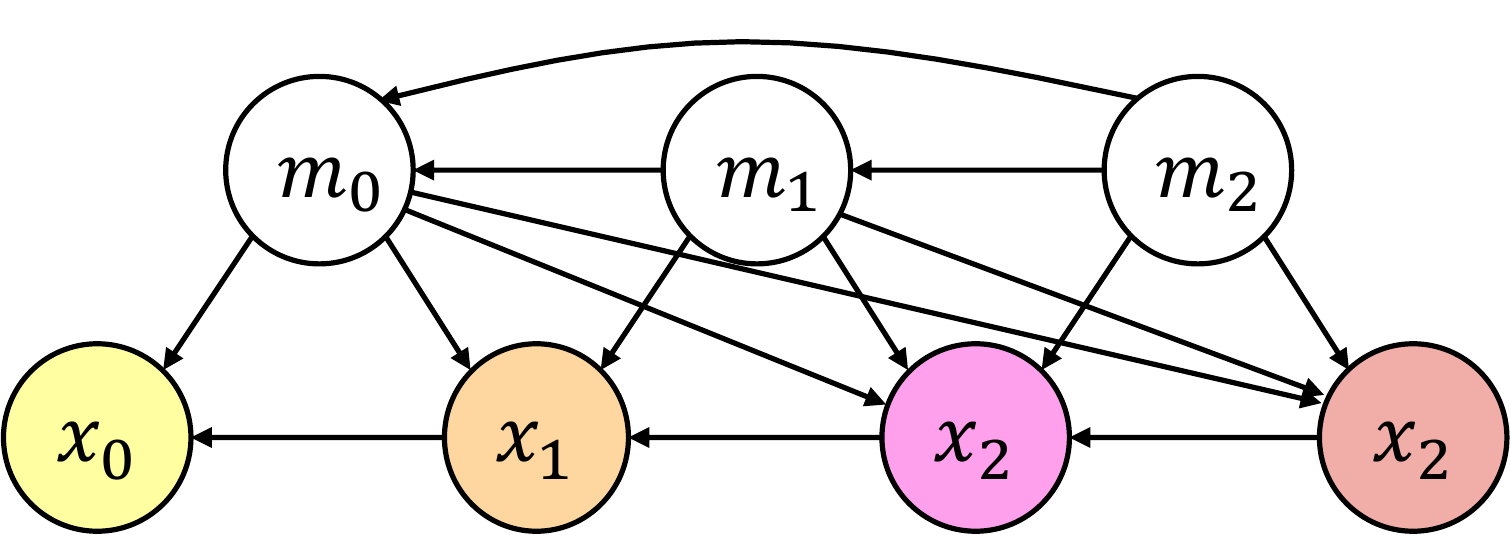}\label{fig:classic}
}

\subfloat[][
Proposed multi-hypothesis smoother tree corresponding to the Bayes net in Figure \ref{fig:classic}. In this case $|\MM| = 2$.
]{
\includegraphics[width=0.79\linewidth]{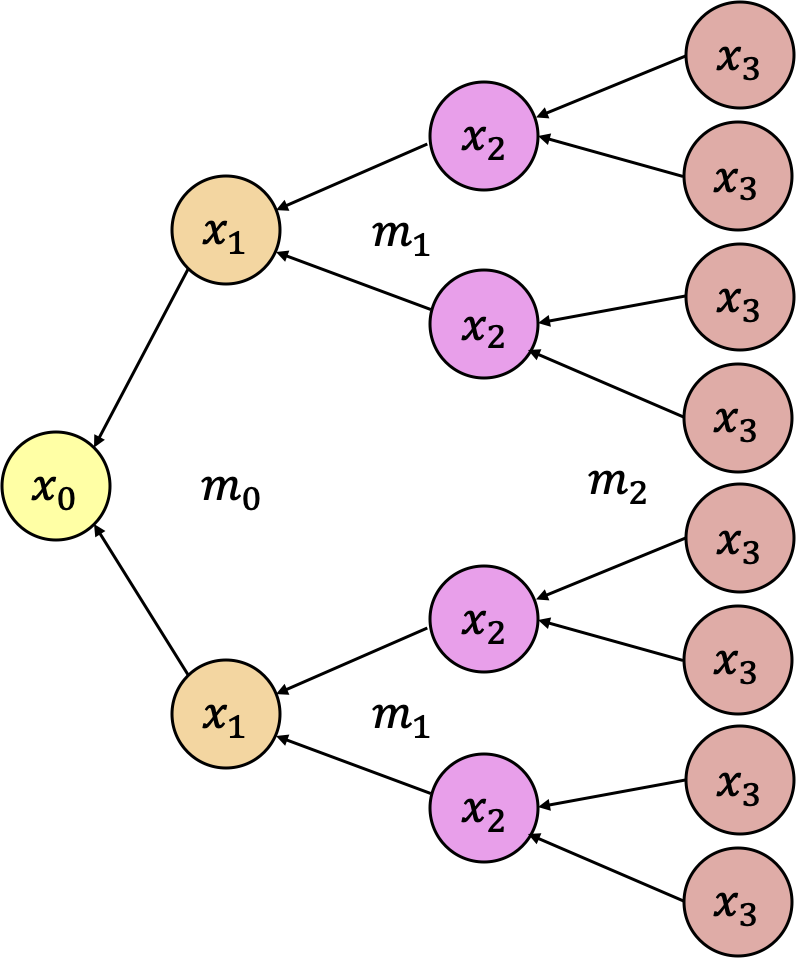}\label{fig:mhs}
}
\caption{Classic Bayes net vs.~tree-structured smoother.}
\label{fig:whammy}
\end{figure}

\begin{figure*}
    \centering
    \includegraphics[width=0.78\textwidth]{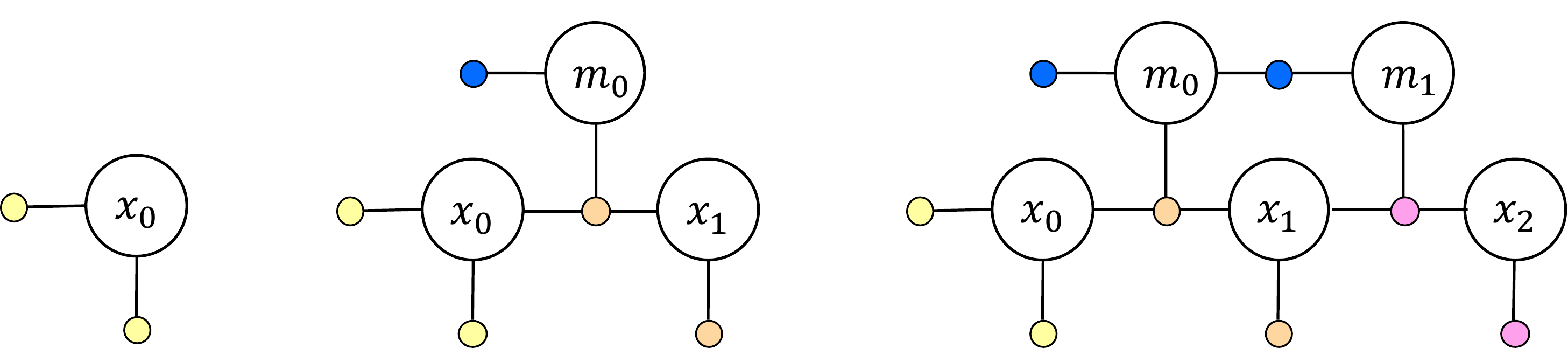}
    \caption{Evolving factor graph during incremental inference.}
    \label{fig:inc-fg}
\end{figure*}

\begin{figure*}
    \centering
    \includegraphics[width=0.78\textwidth]{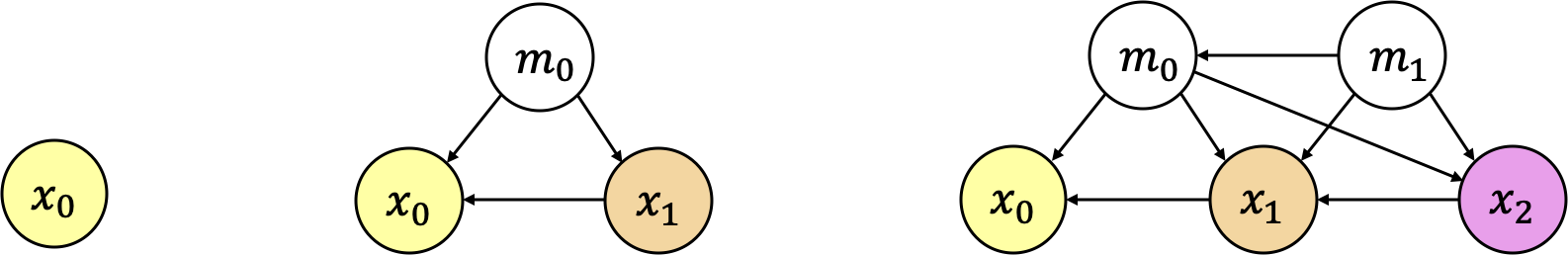}
    \caption{Corresponding Bayes net resulting from eliminating (a).}
    \label{fig:inc-bn}
\end{figure*}

\begin{figure*}
    \centering
    \includegraphics[width=0.78\textwidth]{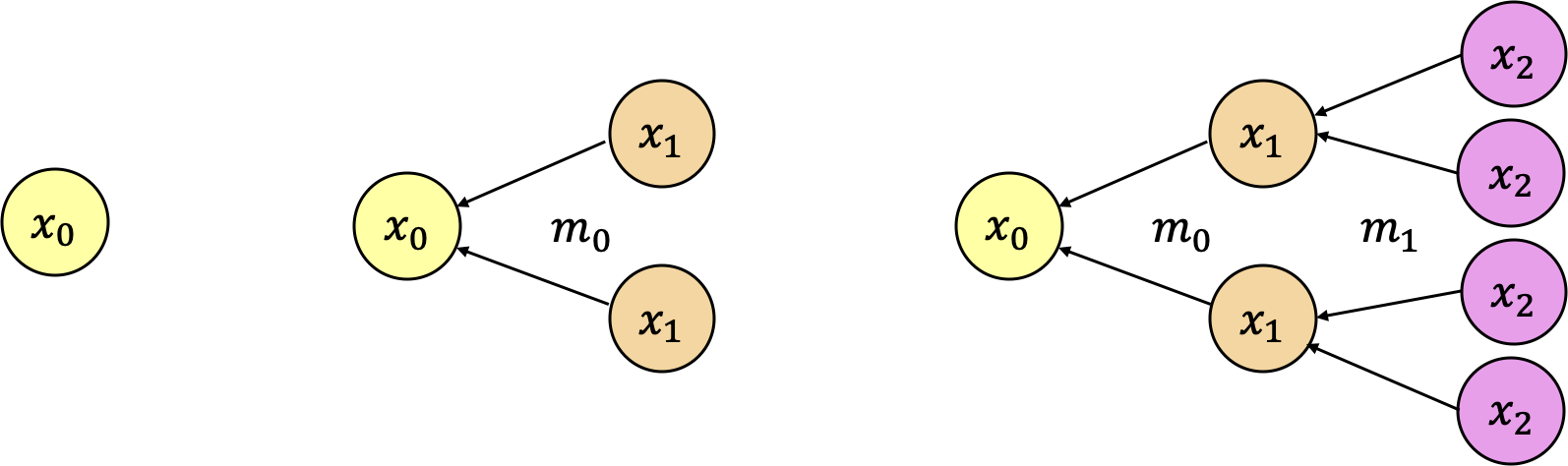}
    \caption{Incremental Multi-hypothesis Smoother over time.}
    \label{fig:inc-mhs}
\end{figure*}

The Bayes net chain \eqref{eq:p_map} can be arranged in a data structure that has the same structure as the hypothesis tree from the MHT~\cite{Reid79tac_MHT}. As discussed above, Equation \ref{eq:p_map} hides $|\MM|^{K-1}$ different chains, one for each mode sequence. A disadvantage of working with that representation is that the arity of each conditional increases over time, and the multi-hypothesis nature of the smoothing problem is hidden from view. However, because each conditional density on $x_{k-1}$ only depends on the subset of modes $M^k\doteq\{m_i\}_0^{k-1}$, we can arrange the conditionals in a tree. As an example, the tree corresponding to Figure \ref{fig:classic} is shown in Figure \ref{fig:mhs}.
 
Formally, the tree distributes every conditional density $p\MAP(x_{k-1}|x_k,M^k)$ in the switching Bayes net \eqref{eq:p_map} over sets of $|\MM|^{k-1}$ \textit{single-mode} switching conditional densities, indexed by their mode history prefix $M^{k-1}$:
\begin{align}
    p\MAP(x_{k-1}|x_k,M^k) \doteq \left\{
    p\MAP_{M^{k-1}}(x_{k-1}|x_k,m_{k-1})
    \right\}_{M^{k-1}}
    \label{eq:sets}
\end{align}
E.g., the first three conditionals correspond to these sets,
\begin{align}
    p\MAP(x_0|x_1,m_0) &\doteq \left\{ 
    p\MAP_{M^0}(x_0|x_1,m_0)
    \right\}_{M^0=\emptyset}
    \nonumber \\
    p\MAP(x_1|x_2,m_0,m_1) &\doteq \left\{
    p\MAP_{M^1}(x_1|x_2,m_1) 
    \right\}_{M^1=0,1}
    \nonumber \\
    p\MAP(x_2|x_3,m_0,m_1,m_2) &\doteq \left\{
    p\MAP_{M^2}(x_2|x_3,m_2) 
    \right\}_{M^2=00,01,10,11}
    \nonumber
\end{align}
which correspond exactly to the inner nodes shown in Figure \ref{fig:mhs}. The final layer in the tree is the set of densities on the final state $\lastx$ conditioned on the entire mode sequence $\Modes$:
\begin{align}
    p\MAP(\lastx|\Modes) \doteq \left\{
    p\MAP_{\Modes}(\lastx)
    \right\}_{\Modes=...}
    \label{eq:leaf_set}
\end{align}

Recovering the optimized state sequence ${\States}\MAP$ for any leaf node $p\MAP_{\Modes}(\lastx)$ can be done simply by back-substitution in reverse time order, following the directed edges in the tree until the root. In particular, the estimate corresponding to the highest value of $\tau(\Modes)$ (Eqn. \ref{eq:p_modes}) corresponds to the MAP estimate. The corresponding optimal mode sequence ${\Modes}\MAP$ can be recovered in a similar fashion, if one takes care to annotate the chosen mode on the edges in the tree.

\subsection{Incremental Inference}
In an incremental scheme this tree grows over time, as illustrated in Figures \ref{fig:inc-fg}-\ref{fig:inc-mhs}.
At each incremental smoothing iteration, all of the leaves of the tree are replicated $|\MM|$ times.
Hence, we have to consider an exponentially growing number of hypotheses at each step. 

However, extending each leaf node towards the future is simple and fast, and we can incrementally keep track of the probability of each mode sequence.
In particular, after $K$ steps, when estimating the next state $\nextx$, assembly and elimination of the factor graph fragment proceeds as in Section \ref{sec:incremental}.
We then use \eqref{eq:p_modes} to calculate $\tau(M^K)$ and \textit{add} it to the previous value $\tau{\Modes}$ to get an updated $\tau$ value.
Book-keeping is simple as each leaf only has to store one $\tau$ value: the number of leaves is exactly $|\MM|^K$ after extending, which is the size of the new $\tau(M^K)$ table.

The integral \eqref{eq:p_modes} depends on the problem, but can be done in closed form in the linear-Gaussian case. In particular, the QR factorization \eqref{eq:QR} yields
\begin{align}
    q\MAP(\nextx) = \exp \left\{ - \frac{1}{2} \| T_m \nextx -  v_m \|^2 - \frac{1}{2} e_m^T e_m \right\} 
\end{align}
where the constant $C_m$ derives from \eqref{eq:f_motion} and \eqref{eq:f_measurement}:
\newcommand{\Imm}{\mathcal{I}_m}
\newcommand{\Izz}{\mathcal{I}_z}
\begin{align}
    C_m = \sqrt{|\Imm\Izz|}~\text{where}~\Imm=Q_m^{-1}~\text{and}~\Izz=R_m^{-1}
\end{align}
Note that $C_m$ can be omitted if the measurements and process covariances are not dependent on the mode $m$.
With this, integration of \eqref{eq:p_modes} for the factor on the \textit{extended} sequence $M^K$ is given by
\begin{align}
    \tau(M^K) = \exp \left\{ -\frac{1}{2} e_m^T e_m \right\} \sqrt{\frac{|\Imm\Izz|}{|T_m^T~T_m|}}
    \label{eq:p_modes2}
\end{align}
The first factor above penalizes error, and the latter is the ratio between the mode-dependent information in the factors (numerator) and the information in the posterior (denominator).
The discrete factor $\tau(M^K)$, of size $|\MM|^K$, can then be multiplied with $p(m_{K-1}|m_{K-2})$ into $p\MAP(\Modes)$ to yield the posterior $p\MAP(M^K)$ on the new mode sequence $M^K$.

Working in log-space avoids numerically unstable multiplication of many small numbers.
Taking the negative log  of \eqref{eq:p_modes2} yields
\begin{align}
    \frac{1}{2} e_m^T e_m + \log |T_m| - \frac{1}{2} \log |\Imm\Izz| 
    \label{eq:p_modes3}
\end{align}
where we made use of the fact that $T_m$ is square.

\subsection{Pruning and Marginalization}
Unbounded computational cost can be avoided by setting a pruning threshold $\theta$, and marginalizing out older states. 
We chose a pruning scheme that evaluates the posterior on the mode sequence $\Modes$ \textit{before} extending the leaves. Any leaf with an associated probability less than $\theta$ will not be extended. In practice we choose $\theta=1\%$ or  $\theta=0.1\%$, which provides a good compromise between accuracy and computation in the examples we tried.

Marginalizing out older states, e.g., after a fixed lag $N$, is straightforward: we simply discard parts of the tree further than $N$ levels from the current time step. Note that this can lead to the tree becoming a forest, but this is not an issue as leaves can continue to be expanded. Another scheme is to simply drop the nodes before the last fork. With aggressive pruning this can occur at a quite shallow depth.

Choosing a high value for $N$ does not incur a high cost. The computational cost of keeping a linear chain before the last fork is zero, and memory load is linear in time. Note however that recovering optimal state sequences at each time does require computation. If this is needed, a wildfire threshold as in iSAM2~\cite{Kaess12ijrr_isam2} can be adopted here as well.

\subsection{Metrics}\label{sec:metrics}
\newcommand{\GT}{Y^{K-1}}
Finally, evaluating estimated mode sequences needs some care.
Each leaf in the tree with parameters $\Theta$ corresponds to a hypothesis for the corresponding mode history. When given a ground truth mode sequence $\GT$, we are faced with evaluating how good our estimate is with respect to it. One metric is to evaluate the likelihood of our model $\Theta$, which is proportional to the probability of producing the ground truth,
\begin{align}
    L(\Theta&; \GT) \propto P(\GT|\Theta) = \nonumber \\
    &\int_{\States} p(\States,\GT|\Measurements) = p\MAP(\GT)
\end{align}
which is exactly the leaf probability evaluated for $\GT$. Typically an error is defined as the negative log of this, i.e.,
\begin{align}
    \mathop{NLL_1}(\Theta;\GT) = - \log p\MAP(\GT).
    \label{eq:NNL}
\end{align}

A more practical metric is to uses the mode marginals. Since we prune hypotheses, and it is likely that the ground truth sequence is not among the unpruned branches, the error \eqref{eq:NNL} is almost always infinite. A more practical error can be obtained by defining the mode marginals
\begin{align}
    p\MAP(M_k=m) = \sum_{\Modes_{k,m}} p\MAP(\Modes)
    \label{eq:marginals}
\end{align}
where $\Modes_{k,m}$ is defined to be the set of all mode sequences having $m_k=m$. The negative log-likelihood of these marginals as a prediction on the ground truth yields the familiar categorical cross entropy loss:
\begin{align}
    \mathop{NLL_2}(\Theta;\GT) &= - \sum_{k} \log p\MAP(M_k=y_k).
    \label{eq:NNL2}
\end{align}

%% file: 5_experiments.tex
Below we demonstrate the the generality of the incremental Multi-Hypothesis Smoother with examples in three problem domains: lane change detection (1D), aircraft maneuver detection (2D), and contact detection in legged robots (3D).

\subsection{Lane Change Detection}

The first example concerns lane change detection in a highway scenario. The trajectories were obtained from the NGSIM Interstate I-80 dataset, which consists of locations of vehicles travelling in a 503m test area on the I-80 highway recorded at 10 Hz. We modeled the lateral dynamics of a car using 2 distinct modes: Constant (CT) and Constant Velocity (CV).
In both cases we use a linear model
\begin{align}
    &p(x_k|x_{k-1}, m) = \mathcal{N}(x_k; F_m x_{k-1}, Q_m)
\end{align}
where the system matrices $F_m$ we use are given by
\begin{align}
    &F_{CT} = \begin{bmatrix}
    1 & 0  \\ 0 & 0
    \end{bmatrix}, 
    F_{CV} = \begin{bmatrix}
    1 & T  \\ 0 & 1
    \end{bmatrix}
\end{align}
with corresponding noise covariance matrices $Q_m$
\begin{align}
    Q_{CT} &= \mathop{diag}(\sigma^2_{CT}, 0) \\
    Q_{CV} &= \mathop{diag}(0, \sigma^2_{CV}).
\end{align}

\begin{figure}
    \centering
    \includegraphics[trim={0 0 0 0},clip,height=5cm]{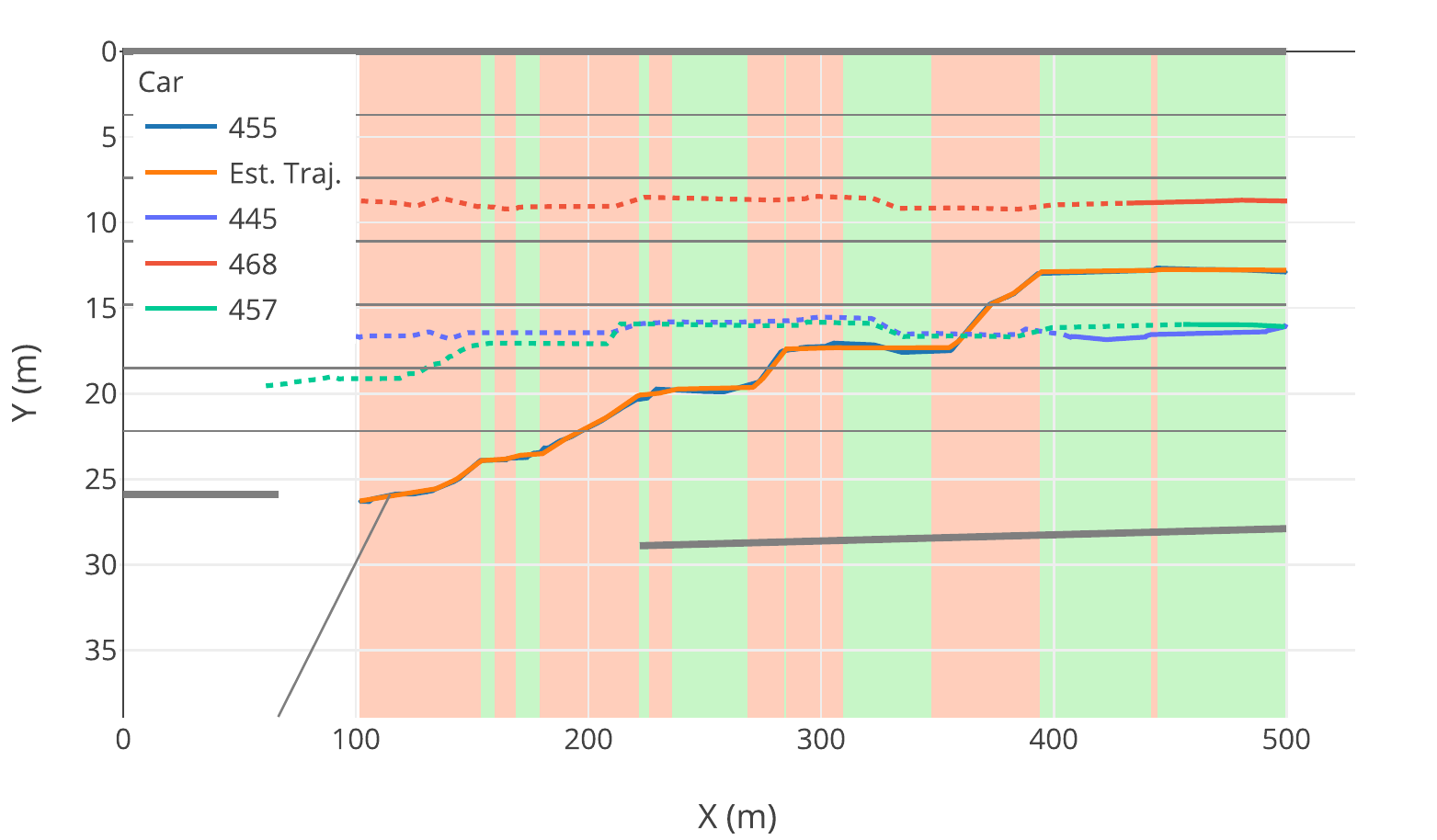}
    \caption{Car trajectory with overlaid mode annotations.\quad Green: CT, Red: CV.}
    \label{fig:fan12_car_lane_change_455}
\end{figure}

Fig. \ref{fig:fan12_car_lane_change_455} shows the detected lane change events for a car driving changing lanes for 4 times in a 500m path.

\subsection{Aircraft Maneuver Detection}
To quantitatively evaluate the iMHS algorithm, we run the algorithm on the 2-mode switching aircraft tracking task from \cite{Hwang06cta_estimation_hybrid}. The aircraft model has 2 discrete modes, constant velocity (CV) and coordinated turn (CT).
The dynamics of this model is given by the following set of discrete-time system equations:
\begin{equation}
    x(k+1) = \begin{bmatrix} 1 & T & 0 & 0 \\ 0 & 1 & 0 & 0 \\ 0 & 0 & 1 & T \\ 0 & 0 & 0 & 1 \end{bmatrix}x(k) + \begin{bmatrix} \frac{T^2}{2} & 0 \\ T & 0 \\ 0 & \frac{T^2}{2} \\ 0 & T\end{bmatrix}u_i(k) + w_i(k)
\end{equation}
\begin{equation}
    y(k) = \begin{bmatrix}1 & 0 & 0 & 0 \\ 0 & 0 & 1 & 0 \end{bmatrix}x(k) + v_i(k)
\end{equation}
In our simulation, $w_i(k)$ and $v_i(k)$ are both zero-mean Gaussian process noise with a standard deviation of $[1,1,1,1]^T$. For each mode, the two constant control inputs were assigned to be $u_{CV}=[0, 0]^T$ and $u_{CT} = [1.5, 1.5]^T$.

\begin{figure}
    \centering
    \includegraphics[width=\linewidth]{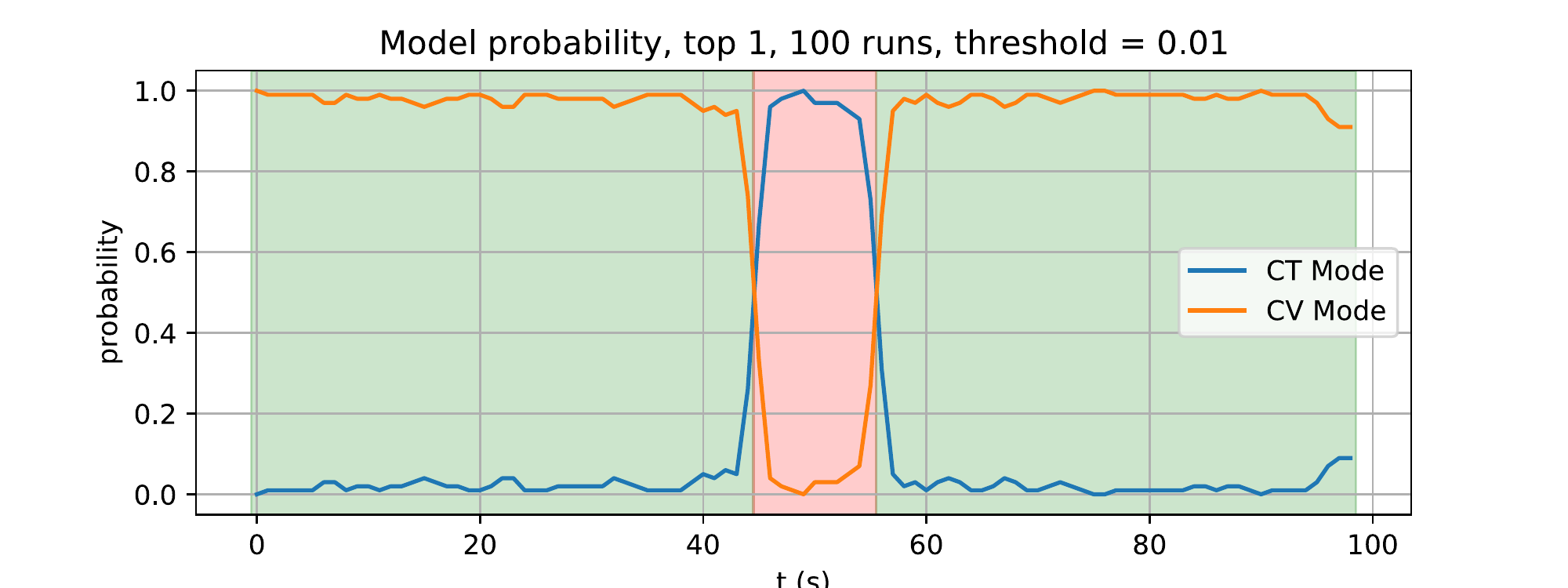}
    \caption{Mode probability of 100 \textit{Monte Carlo} runs of the iMHS algorithm on aircraft example.}
    \label{fig:simple_aircraft_top1}
\end{figure}
The results (Fig. \ref{fig:simple_aircraft_top1}) shows that the iMHS algorithm is able to accurately identify the two modes of flight. In the figure, the green and pink background colors indicate the ground truth mode. Note that the change of mode is identified without any delay, whereas filtering algorithms like RMIMM always have a delay $\geq 1$ time step at state transitions \cite{Hwang06cta_estimation_hybrid}.

\begin{figure}
    \centering
    \includegraphics[width=\linewidth]{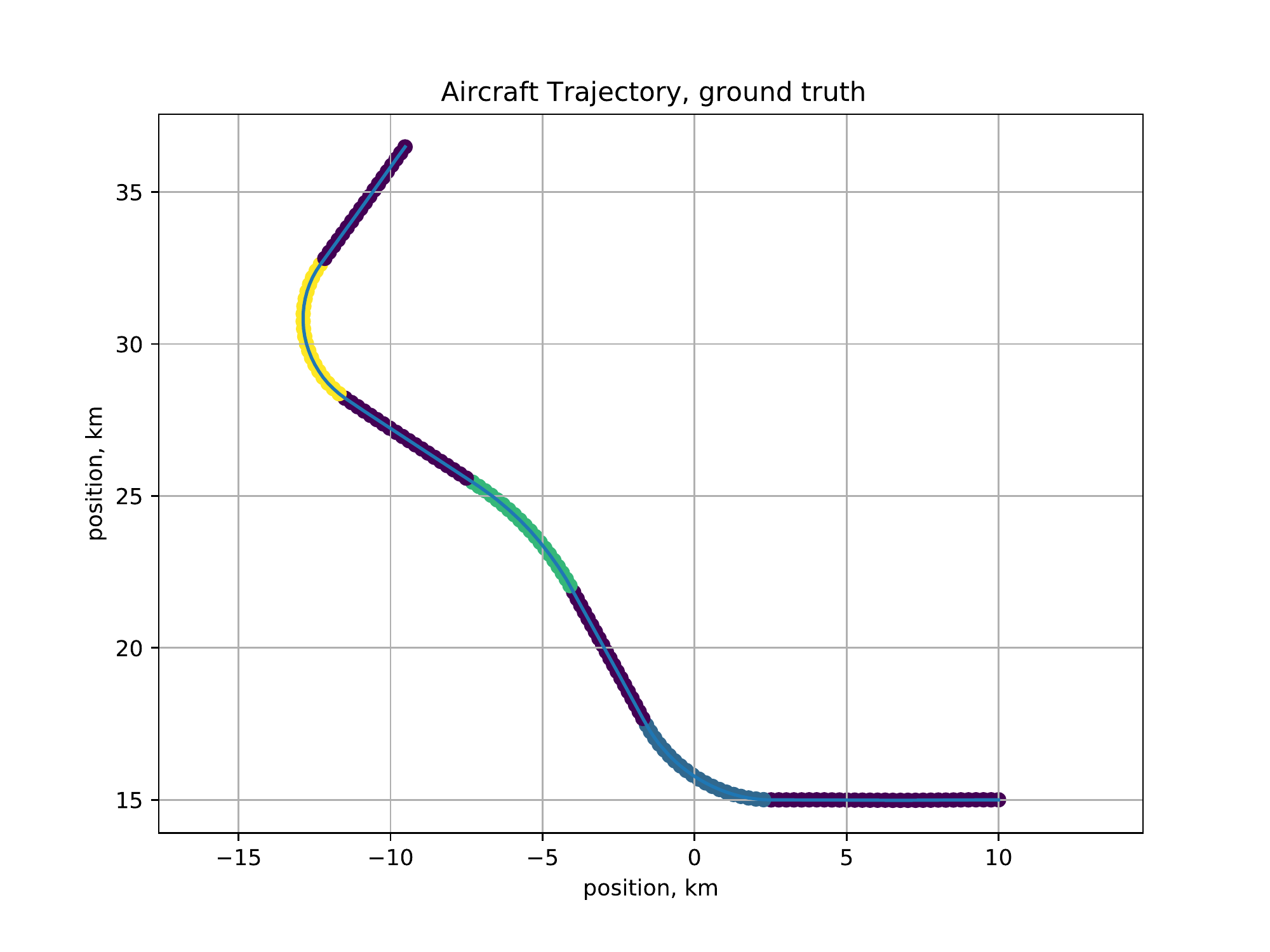}
    \caption{A complex aircraft trajectory with 1 CV mode (blue segments) and 3 CT modes (colored).}
    \label{fig:liu_aircraft_traj}
\end{figure}
To verify the performance of our system we can also run it as a filter. In this mode the algorithm runs online and outputs the current mode estimate as soon as a measurement is taken. We ran this configuration with the more complex trajectory from \cite{Hwang06cta_estimation_hybrid} shown in Fig. \ref{fig:liu_aircraft_traj} which has 7 flight phases: 3 distinct CT modes each with a turn speed of $\omega = -3 m/s$, $\omega = 1.5 m/s$ and $\omega = -4.5 m/s$. 
The plane starts from (x = 10 km, y = 15 km) with an initial speed of 246.93 m/s.

\begin{figure}
    \centering
    \subfloat[Smoother\label{fig:liu_aircraft_top1}]{\includegraphics[width=\linewidth]{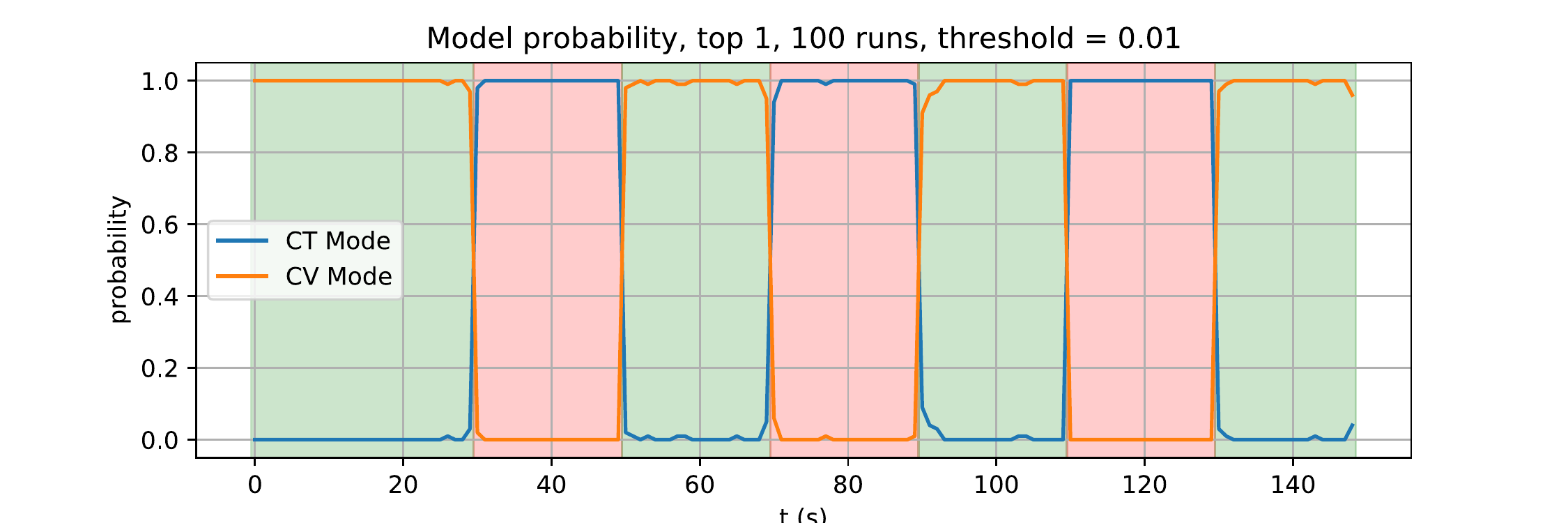}} \quad
    \subfloat[Filter\label{fig:liu_aircraft_top1_filter}]{\includegraphics[width=\linewidth]{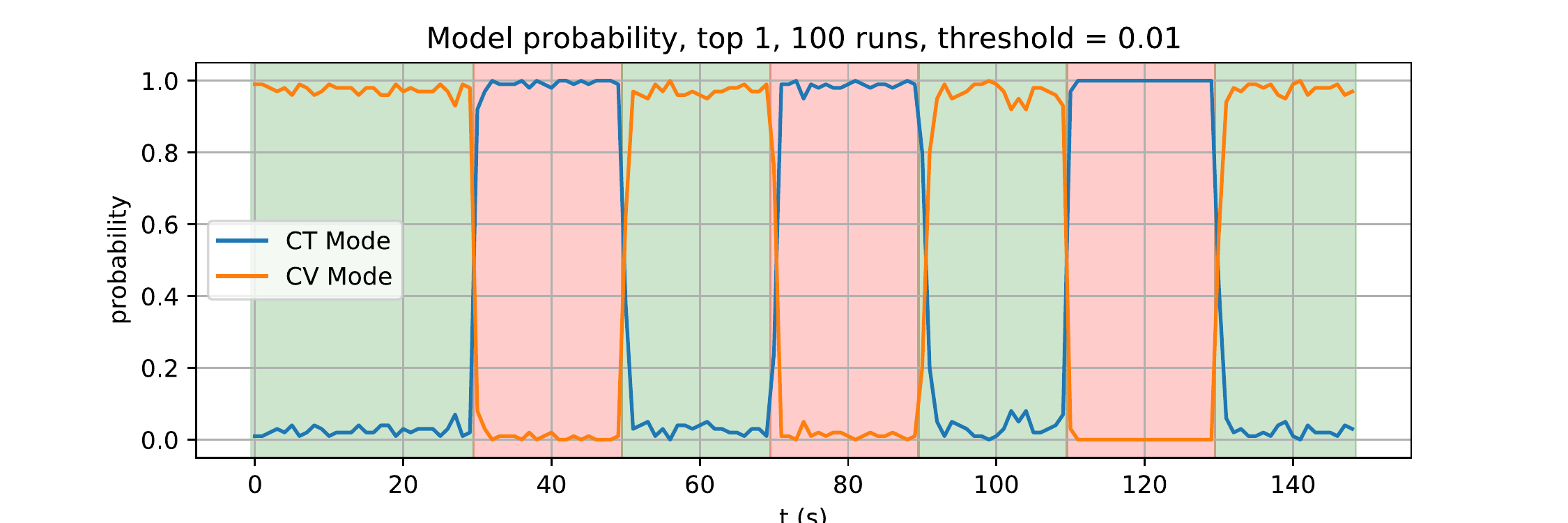}}
    \caption{Mode probability of 100 \textit{Monte Carlo} runs of the iMHS algorithm in smoother mode \textit{v.s.} filter mode (real-time) on a complex trajectory for the flight experiment.}
\end{figure}
Comparison of Fig. \ref{fig:liu_aircraft_top1} and Fig. \ref{fig:liu_aircraft_top1_filter}, show that the iMHS can robustly identify the modes even in the presence of a stochastic process and measurement noise. As expected, in the filter mode the current mode estimate is less accurate than the smoothing mode, due to the high level of noise present in the lateral channel. However, even in filter mode the iMHS algorithm still maintains $>90 \%$ accuracy throughout the trajectory even in filter mode.

\subsection{Contact Estimation in Legged Robots}

\newcommand{\foot}{p^O}
\newcommand{\footB}{p^B}

\begin{figure}
    \centering
    \includegraphics[width=1\linewidth]{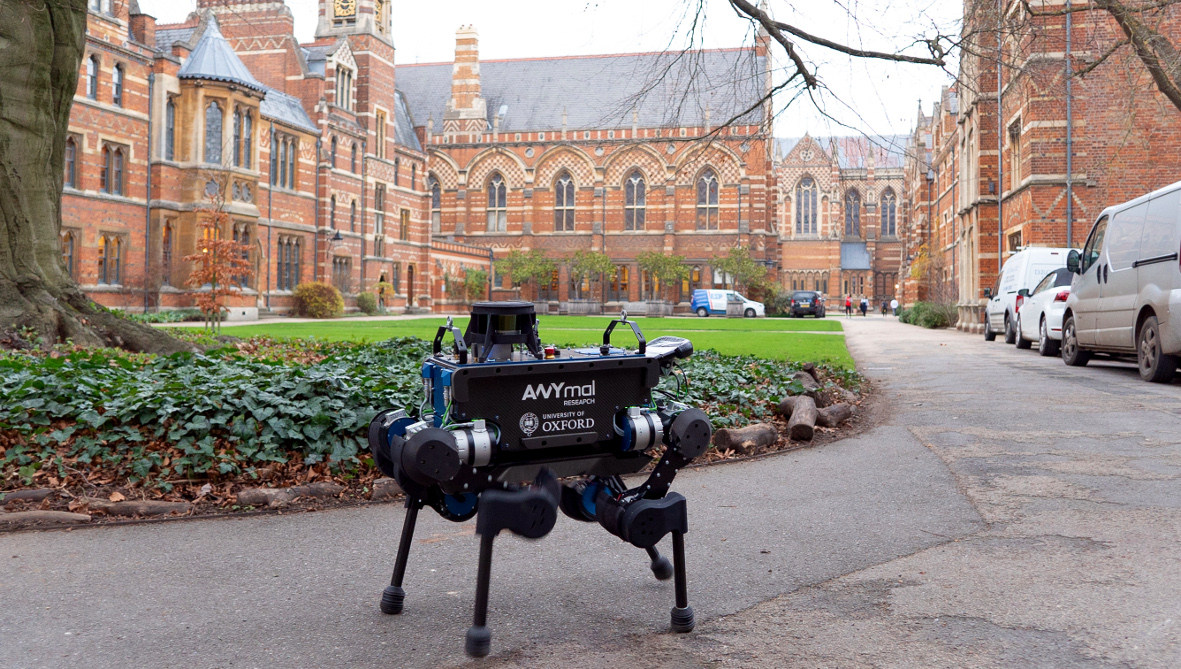}
    \caption{The ANYbotics ANYmal B quadrupedal robot at Keble College, Oxford, United Kingdom.}
    \label{fig:anymal}
\end{figure}
The final application we investigated is inferring the contact state of a quadrupedal robot. We used both simulated data from an A1 and actual log data from an ANYbotics ANYmal quadruped (Fig.~\ref{fig:anymal}) walking in Keble College, Oxford. 
The latter robot and its sensors are described in~\cite{wisth_vilens_2019}, and the dataset we used comprises of several trotting sequences around a courtyard.
The robot's state estimator~\cite{Bloesch12rss} gives us the position of the base link $T^O_B$ and the feet $\foot_i$ in an odometry frame $O$ that is initialized at the start of a run and updated using an internal state estimator. 

We frame the problem as estimating the foot trajectory jointly with the contact sequence $M^{K-1}$ where the mode at time $t_k$ is binary, indicating contact with the ground.
We model the motion of a foot during contact as a 3-dimensional Gaussian density, whose parameters $\mu^O_C$ and $\Sigma_C$ are estimated from training runs. This leads to a quadratic factor between successive foot positions $\foot_{k-1}$ and $\foot_k$
\begin{align}
    &f^C_k(\foot_{k-1},\foot_k) = \|\foot_k - \foot_{k-1} -\mu^O_C\|^2_{\Sigma_C},
    \label{eq:contact}
\end{align}
which is expected to have zero mean with a tight covariance.
During swing, however, we model a constraint on the foot positions $\footB_k$ in the \textit{base} frame, reflecting our knowledge that the feet move rigidly with respect to the body during swing. We obtain:
\begin{align}
    f^S_k(\foot_{k-1},\foot_k) &= \|T^B_{O,k} \foot_k - T^B_{O,k-1} \foot_{k-1} -\mu^B_S\|^2_{\Sigma_S} \nonumber
\end{align}
where we made use of $\footB_k = T^B_{O,k} \foot_k$, with $T^B_O = (T^O_B)^{-1}$. 
We have
$T^B_O \foot = T^R_O \foot + t^B_O$
and we assume $R^B_{O,k} \approx R^B_{O,k-1}$, which allows us to simplify this into another quadratic factor,
\begin{align}
    f^S_k(\foot_{k-1},\foot_k) &= 
    \|\foot_k  -\foot_{k-1} - (\Delta t^O_{B,k-1,k} + R^O_{B,k} \mu^B_S)
    \|^2_{\Sigma_S}. \nonumber
\end{align}
where $\Delta t^O_{B,k-1,k}=t^O_{B,k} - t^O_{B,k-1}$.


\begin{figure}
    \centering
    \subfloat[Gait Marginal Probabilities\label{fig:A1a}]
    {\includegraphics[trim={2cm 7cm 2cm 7cm},clip,width=\linewidth]{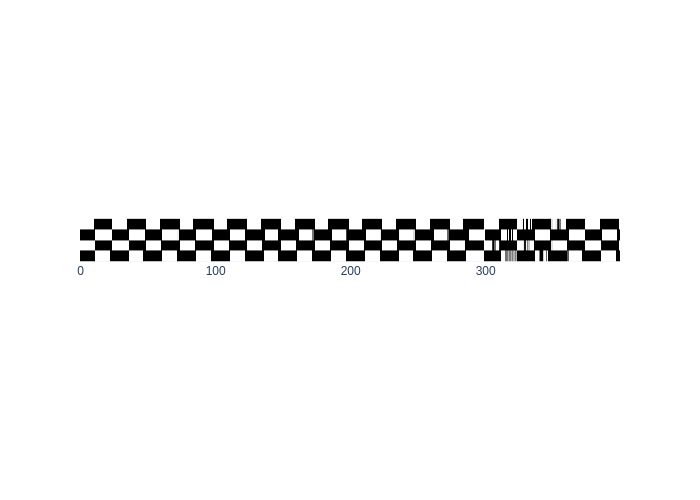}}
    
    \subfloat[Gait MAP Estimate\label{fig:A1b}]
    {\includegraphics[trim={2cm 7cm 2cm 7cm},clip,width=\linewidth]{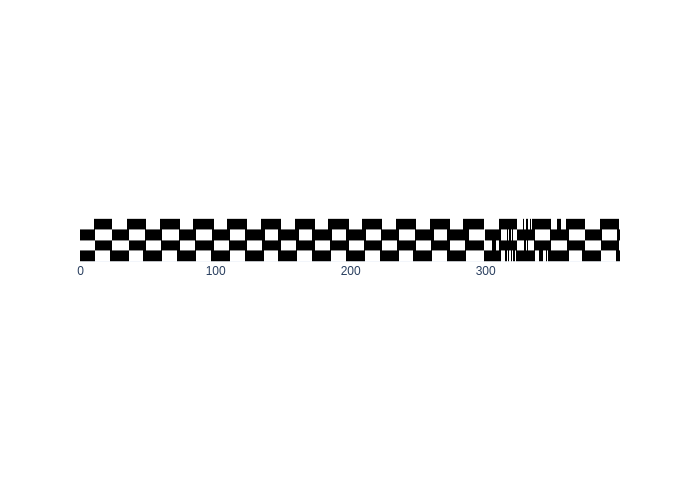}}
    
    \subfloat[Gait Error\label{fig:A1c}]
    {\includegraphics[trim={2cm 7cm 2cm 7cm},clip,width=\linewidth]{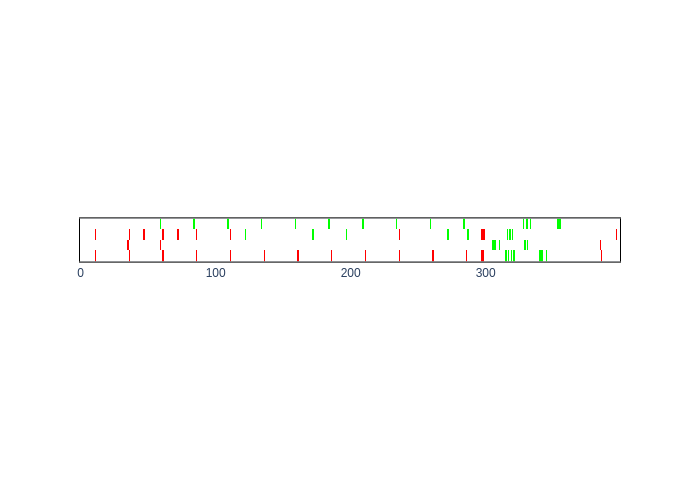}} \quad
    
    \caption{A1 simulation results. Panel (a) and (b) represents the marginal probabilities \eqref{eq:marginals} and the MAP estimate for the contact sequence of all feet, with black signifying contact. Panel (c) is the error for the MAP estimate, color-coded with red for incorrectly predicted contact, green for incorrectly predicted flight, and white for no error.} 
    \label{fig:A1}
\end{figure}


\begin{figure}
    \centering
    \subfloat[Gait Marginal Probabilities\label{fig:Keblea}]
    {\includegraphics[trim={2cm 7cm 2cm 7cm},clip,width=\linewidth]{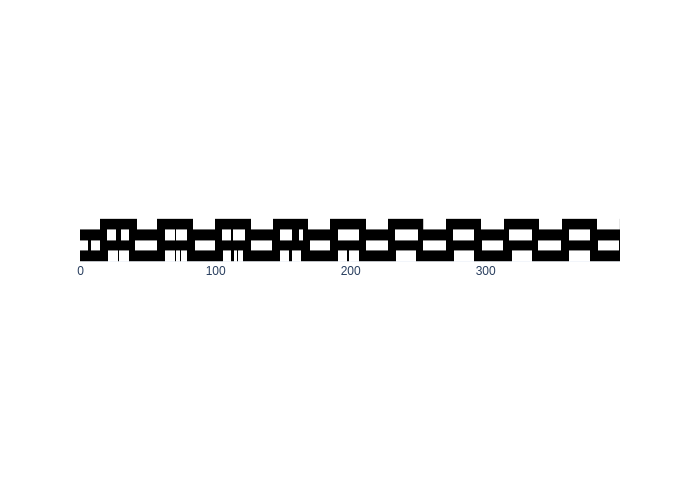}}
    
    \subfloat[Gait MAP Estimate\label{fig:Kebleb}]
    {\includegraphics[trim={2cm 7cm 2cm 7cm},clip,width=\linewidth]{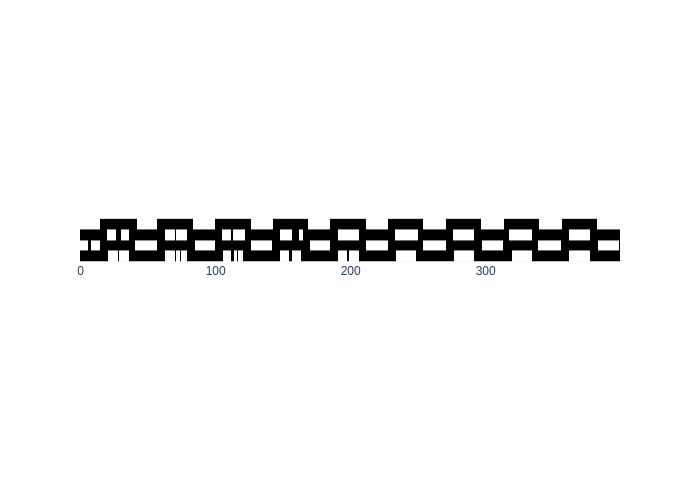}}
    
    \subfloat[Gait Error\label{fig:Keblec}]
    {\includegraphics[trim={2cm 7cm 2cm 7cm},clip,width=\linewidth]{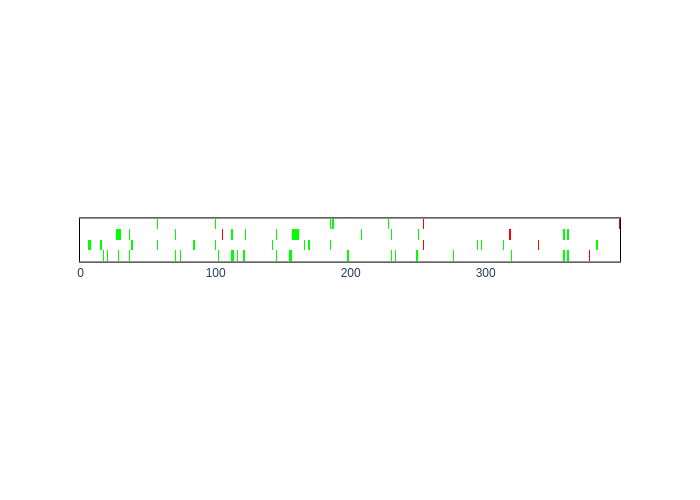}} \quad
    
    \caption{ANYmal B. See caption of Figure \ref{fig:Keble}}
    \label{fig:Keble}
\end{figure}

\begin{table}
\centering
 \begin{tabular}{| c | c | c | c | c |} 
 \hline
 \textbf{Platform} & \textbf{LH} & \textbf{LF} & \textbf{RF} & \textbf{RH} \\
 \hline\hline
A1 Simulation & 1.013 & 1.089 & 0.377 & 1.242 \\ 
 \hline
 ANYmal Data & 0.819 & 0.639 & 1.362 & 1.284 \\
 \hline
\end{tabular}
\caption{Cross-entropy error. This measures the difference between our estimated contact states and the pseudo-ground-truth states. Lower values indicate better performance.}
\label{tab:ce}
\end{table}
Qualitative results for the A1 simulation and the Keble dataset are shown in Figures~\ref{fig:A1} and~\ref{fig:Keble}, respectively.
For quantitative evaluation we use the contact states provided by the probabilistic contact detector from~\cite{Jenelten2019} as pseudo-ground-truth. 
As described in Section \ref{sec:metrics} we report on the categorical cross-entropy error between the contact state marginals estimated by the iMHS and the ground truth contact states in Table \ref{tab:ce}.

%% file: 6_conclusions.tex
The incremental Multi-Hypothesis Smoother proposed in this paper provides a solid foundation for inference over time in hybrid dynamical systems. Our results above demonstrate the applicability in various domains, but we are most excited about the possibility of integrating this with a state of the art inertial state estimation pipeline. This would enable state estimation in a variety of legged robots without the need for contact sensors or dedicated force sensors. Smoothing rather than filtering will enable state estimators to more accurately update continuous states that depend on hybrid states. 